\documentclass{article}
\usepackage{spconf,amsmath,graphicx}
\usepackage{amssymb}
\usepackage{epstopdf}
\usepackage{booktabs}
\usepackage{makecell}

\usepackage[table]{xcolor} % 加載 xcolor 套件，支持表格上色
\usepackage{booktabs} % 提供更好的表格格式
\usepackage{graphicx} % 支持 resizebox
\usepackage{paralist}
\usepackage[pagebackref,breaklinks,colorlinks]{hyperref}

\newcommand{\heading}[1]{
\vspace{1mm}\noindent\textbf{#1}}

\title{BEVANET: BILATERAL EFFICIENT VISUAL ATTENTION NETWORK FOR REAL-TIME SEMANTIC SEGMENTATION}
\name{Ping-Mao Huang$^{\star}$ \qquad I-Tien Chao$^{\star}$ \qquad Ping-Chia Huang$^{\dagger}$ \qquad Jia-Wei Liao$^{\dagger}$ \qquad Yung-Yu Chuang$^{\dagger}$
\thanks{\hspace*{-1.8em}{\footnotesize{Thanks to NSTC, ASUSTek, and NTU for their support through grants 113-2634-F-002-007, 113-2221-E-002-112-MY3, 113L9009 and 114L892203.}}}
}

\address{$^{\star}$Graduate Institute of Networking and Multimedia, National Taiwan University \\ $^{\dagger}$Department of Computer Science and Information Engineering, National Taiwan University}

\begin{document}
%\ninept
%
\maketitle
%
% \input{front/abstract template}
% \input{front/keywords template}
% % Contents of Thesis
% \input{contents/Introduction Template}
% \input{contents/Format}
% \input{contents/Pagestyle}
% \input{contents/Typestyle} 
% \input{contents/Major Head} 
% \input{contents/Print}
% \input{contents/Page}
% \input{contents/Illustrations}
% \input{contents/Footnotes}

% !TeX root = ../main.tex

\begin{abstract}
Real-time semantic segmentation presents the dual challenge of designing efficient architectures that capture large receptive fields for semantic understanding while also refining detailed contours. Vision transformers model long-range dependencies effectively but incur high computational cost. To address these challenges, we introduce the Large Kernel Attention (LKA) mechanism. Our proposed Bilateral Efficient Visual Attention Network (BEVANet) expands the receptive field to capture contextual information and extracts visual and structural features using Sparse Decomposed Large Separable Kernel Attentions (SDLSKA). The Comprehensive Kernel Selection (CKS) mechanism dynamically adapts the receptive field to further enhance performance. Furthermore, the Deep Large Kernel Pyramid Pooling Module (DLKPPM) enriches contextual features  by synergistically combining dilated convolutions and large kernel attention. The bilateral architecture facilitates frequent branch communication, and the Boundary Guided Adaptive Fusion (BGAF) module enhances boundary delineation by integrating spatial and semantic features under boundary guidance. BEVANet achieves real-time segmentation at 33 FPS, yielding 79.3\% mIoU without pretraining and 81.0\% mIoU on Cityscapes after ImageNet pretraining, demonstrating state-of-the-art performance. The code and model is available
at \url{https://github.com/maomao0819/BEVANet}.
%Real-time semantic segmentation development encounters challenges in creating efficient architectures that capture large receptive fields for semantic comprehension and detailed contour refinement.
%Vision transformers excel at capturing long-range dependencies; however, they are inefficient. To overcome these challenges, we introduce the Large Kernel Attention mechanism into this area. We propose the Bilateral Efficient Visual Attention Network (BEVAN), which enlarges the receptive field to capture contextual information and extracts visual and structural features using Sparse Decomposed Large Separable Kernel Attentions (SDLSKA). The Comprehensive Kernel Selection (CKS) mechanism dynamically adapts the receptive field for improved performance. Additionally, the Deep Large Kernel Pyramid Pooling Module (DLKPPM) enriches contextual features through dilated convolution and large kernel attention. The bilateral architecture facilitates frequent communication between branches, while the Boundary Guided Attention Fusion (BGAF) module enhances boundary delineation by merging spatial and semantic features with boundary guidance. Our model achieves real-time segmentation at 30 FPS, attaining 79\% mIoU without pretraining and 81\% mIoU on Cityscapes after ImageNet pretraining, demonstrating state-of-the-art performance.
\end{abstract}
% !TeX root = ../main.tex

% Contents of Thesis
% !TeX root = ../main.tex

\section{Introduction}
\label{sec:intro}

% Semantic segmentation is crucial in computer vision, assigning class labels to each pixel in an image. It requires precise boundary detection, semantic context comprehension, and object completeness, making it a dense prediction task aimed at improving performance and efficiency. Traditional segmentation relied on hand-crafted features. Since Fully Convolutional Networks (FCN) \cite{long2015fully} introduced end-to-end dense prediction, models like UNet \cite{ronneberger2015u} with skip connections and encoder-decoder structures have improved segmentation. PSPNet \cite{zhao2017pyramid} significantly improved performance using pyramid pooling. Despite advancements, many models face computational complexity, limiting real-time applications in areas like autonomous driving and robotics.

Semantic segmentation, assigning class labels to every pixel, is vital in computer vision. Initially reliant on hand-crafted features, it evolved with Fully Convolutional Networks~\cite{long2015fully} and models like UNet~\cite{ronneberger2015u}, improving performance with encoder-decoder structures and skip connections. PSPNet~\cite{zhao2017pyramid} enhanced results using pyramid pooling. 
However, computational demands of many models render them impractical for real-time applications like autonomous driving and robotics.

% Semantic segmentation assigns class labels to every pixel in an image and is critical for applications like autonomous driving and robotics. Early models relied on handcrafted features, while modern approaches leverage Fully Convolutional Networks \cite{long2015fully} and models like UNet \cite{ronneberger2015u}, improving performance with encoder-decoder structures and skip connections. PSPNet \cite{zhao2017pyramid} enhanced results using pyramid pooling. Despite these advances, achieving real-time performance remains challenging due to computational constraints, particularly in capturing fine details and semantic context.

\begin{figure}[!ht]
\centering
\includegraphics[width=\linewidth]{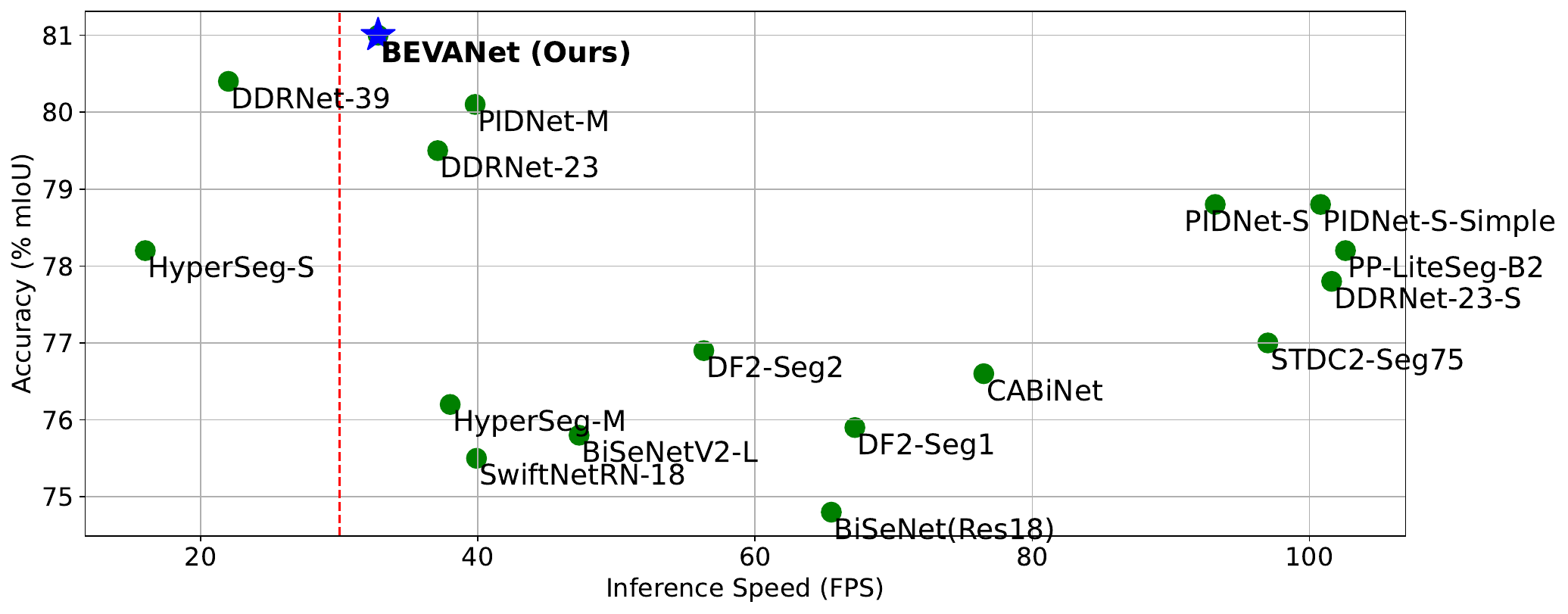}
\caption{Performance of real-time models on the Cityscapes~\cite{cordts2016cityscapes} validation set, with our model in blue and others in green.}
%\vspace{-2.5mm}
\label{fig:plot}
\end{figure}

% Real-time semantic segmentation seeks to balance speed and accuracy. Lightweight architectures use depth-wise separable convolutions to improve efficiency. Recent designs like BiSeNet \cite{yu2021bisenet}, STDC \cite{fan2021rethinking}, and DDRNet \cite{hong2021deep} combine low-level spatial features with high-level semantic context for promising performance. PIDNet \cite{xu2023pidnet} achieves state-of-the-art results by applying Proportional-Integral-Derivative controller principles, incorporating three branches to process detailed, contextual, and boundary information. However, challenges remain, such as insufficient receptive fields for contours.

Real-time semantic segmentation seeks to balance speed and accuracy through lightweight architectures, often leveraging depth-wise separable convolutions. Recent models like BiSeNet \cite{yu2021bisenet}, STDC \cite{fan2021rethinking}, and DDRNet \cite{hong2021deep} blend spatial and semantic features to improve efficiency. PIDNet \cite{xu2023pidnet} achieves state-of-the-art results by applying Proportional-Integral-Derivative controller principles, incorporating three branches to process detailed, contextual, and boundary information. However, there are insufficient receptive fields for contours.

The Vision Transformer \cite{dosovitskiy2020image} and its variants introduced superior long-range dependency modeling but are computationally heavy. Efficient attention mechanisms such as SeaFormer \cite{wan2023seaformer} attempt to optimize this, but challenges remain.

% Vision Transformer (ViT) \cite{dosovitskiy2020image} and it variant revolutionized the field with superior long-range dependency modeling but require high computational resources. Efficient attention mechanisms in SeaFormer \cite{wan2023seaformer} attempt to optimize this, but challenges persist.

Recent approaches like RepLKNet \cite{ding2022scaling} highlight the advantages of Large Kernel Attention (LKA), integrating convolution and attention mechanisms to capture global context effectively. SLaK~\cite{liu2022more} expanded kernel sizes to 51 by replacing a large kernel with two long parallel kernels and a small kernel. VAN~\cite{guo2023visual} and LSKA~\cite{lau2024large} further optimized this with dilated and strip convolution, reducing computational demands. LSKNet~\cite{li2023large} introduced the selective kernel concept from SKNet~\cite{li2019selective}. However, these approaches still lack multi-scale feature integration and receptive field adjustment.

We introduce the LKA mechanism into our Bilateral Efficient Visual Attention Network (BEVANet) to address challenges in real-time semantic segmentation, such as capturing contour details, semantic context, and fusing features at different levels. 
Our design features the Efficient Visual Attention (EVA) block with Sparse Decomposed Large Separable Kernel Attention (SDLSKA) to expand the receptive field, capture semantic context, and refine details. We design the Comprehensive Kernel Selection (CKS) mechanism, which integrates features from both small and large kernels using dynamic channel and spatial attention. Additionally, we propose the Deep Large Kernel Pyramid Pooling Module (DLKPPM) to enrich contextual features and prevent information loss typically caused by pooling and striding in traditional methods. 
We also develop a bilateral architecture that facilitates continuous communication between two branches and the Boundary Guided Attention Fusion (BGAF) module that adapts semantic and detail fusion with boundary information. The interactions of these branches improve the representation of features by integrating various features. 
As shown in Fig.~\ref{fig:plot}, BEVANet offers a robust and efficient framework that achieves state-of-the-art real-time segmentation by effectively balancing accuracy and efficiency.
Main contributions are listed as follows.
\begin{compactitem}
\item \textbf{Efficient Attention Mechanisms.} We leverage large kernel attention to design the EVA block, SDLSKA, CKS, and DLKPPM modules. These components enlarge and dynamically adjust receptive fields, enhance feature representation, capture contextual information, and refine object details, improving spatial modeling.
\item \textbf{Branch Interaction.} Frequent communication between high- and low-level branches through the bilateral architecture and the BGAF module enhances semantic concepts and detail contour by sharing information and shortcut, enabling adaptive feature fusion.
\item \textbf{Performance.} 
% BEVANet balances speed and accuracy better than existing models. It achieves real-time segmentation over 30 FPS with 81.0\% mIoU on Cityscapes after ImageNet pre-training and maintains 79.3\% mIoU without pre-training, showing reduced dependency on large pre-training datasets. Its variants, BEVANet-S, further achieves 83\% on CamVid, demonstrating scalability.
BEVANet offers a superior balance of speed and accuracy to existing models. It achieves real-time segmentation over 30 FPS with 81.0\% mIoU on Cityscapes after ImageNet pre-training and  79.3\% without, indicating reduced dependency on large pre-training datasets. Its variant BEVANet-S further achieves 83\% mIoU on CamVid, demonstrating its scalability.
\end{compactitem}

% !TeX root = ../main.tex

\section{Methodology}
\label{sec:method}

\begin{figure}[!htb]
\begin{minipage}[b]{\linewidth}
  \centering
  \centerline{\includegraphics[width=\linewidth]{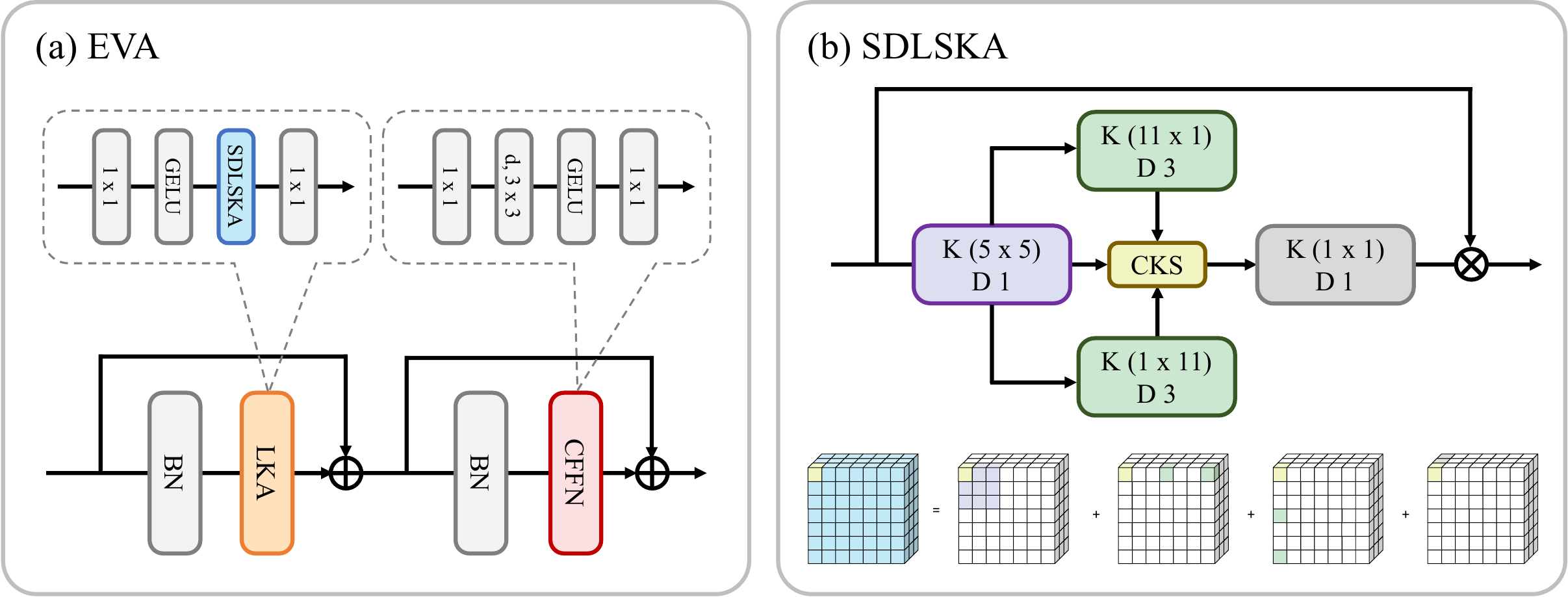}}
  \centerline{\textbf{The structure of (a) EVA block and (b) SDLSKA module.}}\smallskip
\end{minipage}
\hfill
\begin{minipage}[b]{\linewidth}
  \centering
  \centerline{\includegraphics[width=\linewidth]{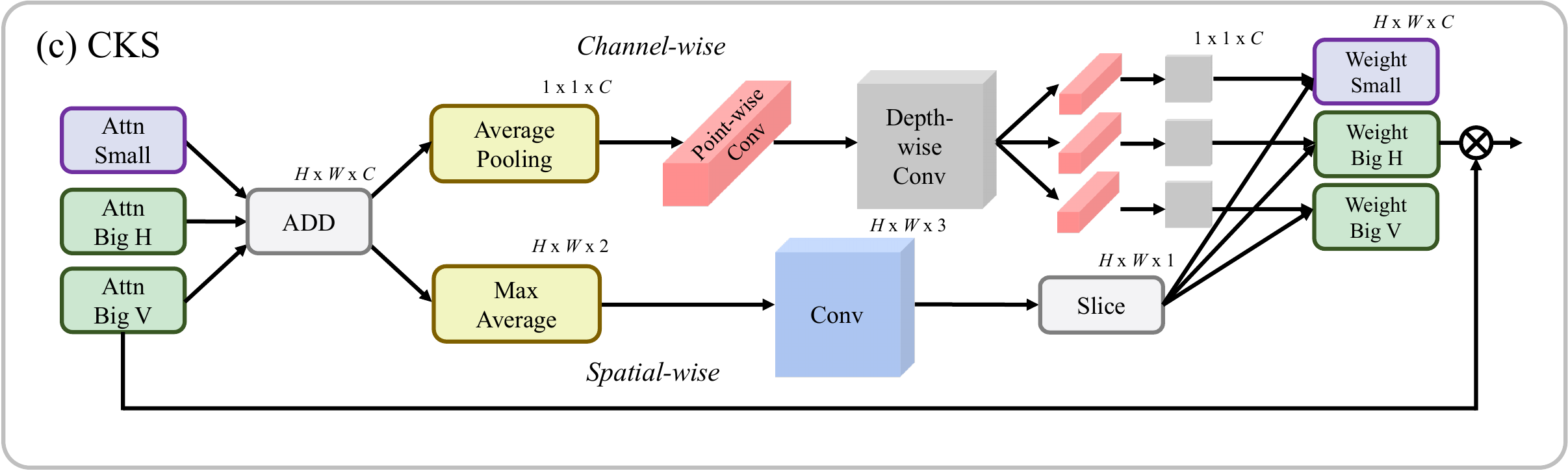}}
  \centerline{(c) \textbf{The structure of CKS module.}}
\end{minipage}
\caption{Our EVA block (a) utilizes the SDLSKA (b) and CKS (c) modules to adaptively expand the vision. The SDLSKA module extracts large features using a $5 \times 5$ convolution, followed by $1 \times 11$ and $11 \times 1$ strip convolutions with a dilation rate of 3, achieving a receptive field of 35. These features are fused with the CKS mechanism and refined via a point-wise convolution. The CKS module computes weights across channel and spatial dimensions through pooling and aggregation. In the spatial branch, a convolution operation generates three feature channels. The channel branch refines features using pointwise and depthwise convolution. The fusion of spatial and channel weights, facilitated by the multiplication.}
\label{fig:EVA}
\end{figure}

We propose the Efficient Visual Attention (EVA) module, utilizing Sparse Decomposed Large Separable Kernel Attentions (SDLSKA) and Comprehensive Kernel Selection (CKS) to adaptively enlarge the receptive field.
The Deep Large Kernel Pyramid Pooling Module (DLKPPM) leverages large kernels for contextual enrichment. Additionally, the Bilateral Architecture (BA) and Boundary Guided Adaptive Fusion (BGAF) facilitate feature interaction across two branches.

\subsection{Bilateral Architecture}
\label{ssec:BA}

\begin{figure*}[!ht]
\centering
\includegraphics[width=0.8\linewidth]{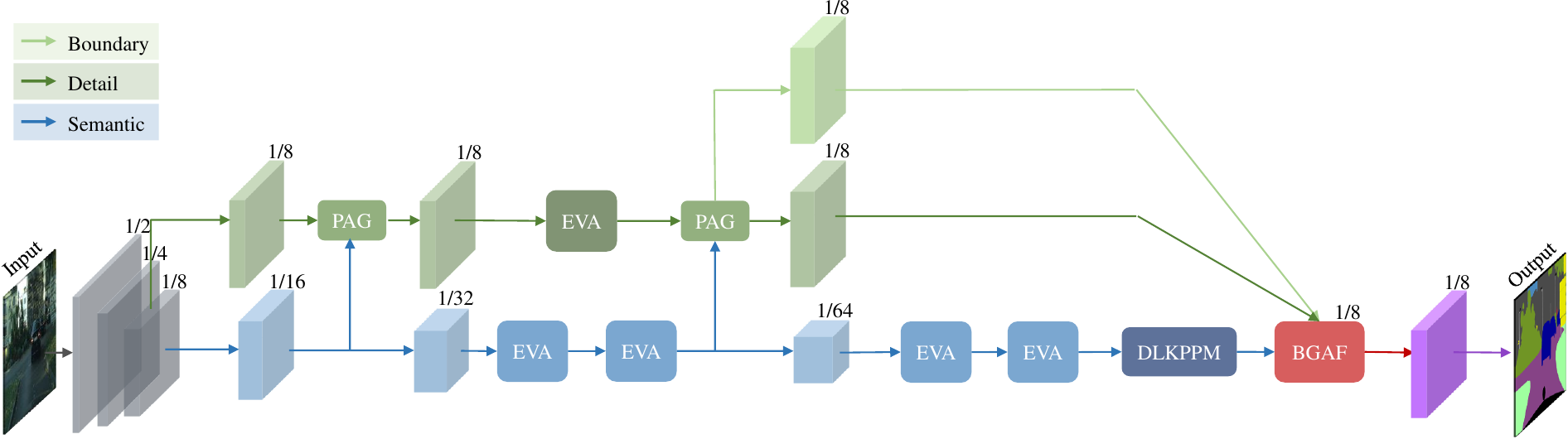}
\caption{\textbf{The overall structure of the BEVANet.} The high-level branch captures contextual information and long-range dependencies, enriching the spatial branches with semantics. Meanwhile, the low-level branch preserves high-resolution details and high-frequency features for accurate boundaries, maintaining a resolution of 1/8 of the original size to retain fine contours.}
\label{fig:BA}
\end{figure*}

Inspired by PIDNet~\cite{xu2023pidnet}, we propose BA, which frequently integrates semantic context into detailed features. As depicted in Fig.~\ref{fig:BA}, it features a high-level branch to capture semantic concepts by reducing feature maps and a low-level branch to extract contours and detect boundaries while maintaining resolution. This enables continuous branch interaction to improve semantic understanding and object boundary .

% Inspired by PIDNet [8], we propose BA, which frequently integrates semantic context into detailed features. As depicted in Fig. 3, it features a high-level branch for capturing semantic concepts with reduced feature maps, and a low-level branch for extracting contours and detecting boundaries while maintaining resolution. This design enables continuous branch interaction to improve semantic understanding and object boundary delineation.

% Inspired by PIDNet \cite{xu2023pidnet}, BEVAN’s bilateral design integrates a high-level branch for semantic context and a low-level branch for boundary refinement. Continuous interaction between branches ensures precise feature fusion.

\subsection{Efficient Visual Attention Block}
\label{ssec:EVA}

Our EVA Block is inspired by the robust block design in VAN \cite{guo2023visual} and LSKA \cite{lau2024large}, demonstrated in Fig.~\ref{fig:EVA}. It consists of two main components: the Large Kernel Attention (LKA) and the Convolution Feed-Forward Network (CFFN).

The LKA block captures long-range dependencies. By utilizing SDLSKA, it expands the receptive field to a broader context, resulting in more effective semantic capture. It also uses the CKS mechanism to combine features from kernels of different sizes and adjusts the vision accordingly. This synergistic combination renders EVA Block an efficient module for robust feature extraction and precise feature representation.

The CFFN block refines and integrates features to ensure a well-balanced and informative output. In addition, it refines features channel-wise using pointwise convolution.

\subsubsection{Sparse Decompose Large Separable Kernel Attentions}
\label{sssec:SDLSKA}

The SDLSKA module is designed to expand the receptive field to effectively capture semantic information and refine details. The structure is depicted in Fig.~\ref{fig:EVA}(b). Drawing from SLaK~\cite{liu2022more}, we simplify large-kernel computation through sparse grouping by decomposing them into a smaller convolution and two strip dilation kernels, then adaptively fusing them using CKS module. The smaller convolution helps focus on specific areas, while the two strip dilation kernels refine the focus, with low computation. Additionally, inspired by LSKA~\cite{lau2024large}, we combine strip convolutions with depthwise, pointwise, and dilated convolutions to capture large-kernel features efficiently. This approach reduces parameters while leveraging 2D structural information, resulting in better computational efficiency. It also adapts effectively to spatial and channel dimensions to capture long-range dependencies.

\vspace{-2mm}
\subsubsection{Comprehensive Kernel Selection}
\label{sssec:CKS}

Our CKS module innovatively performs joint channel-wise and spatial-wise adjustments, crucial for the receptive field adaptation of SDLSKA and for dynamic multi-scale feature fusion from kernels of varying shapes. This integration surpasses SKNet~\cite{li2019selective} and LSKNet~\cite{li2023large}, which address these dimensions in a decoupled manner.
This approach captures the interdependence between channels and spatial dimensions, which is crucial for effective feature integration. As shown in Fig.~\ref{fig:EVA}(c), the module efficiently manages complex fusion across these dimensions. By merging features across kernel scales and considering interdependencies, it enables a holistic, adaptable fusion that leads to richer representations.

%Our CKS module dynamically adjusts the receptive field of SDLSKA and fuses multi-scale features of varying shapes by considering both channel-wise and spatial-wise adjustments, in contrast to SKNet \cite{li2019selective} and LSKNet \cite{li2023large}, which handle these aspects separately.
%Our method addresses the interdependence between dimensions, crucial for effective feature integration. As shown in Fig.~\ref{fig:EVA}(c), the module efficiently manages complex fusion across spatial and channel dimensions, enhancing performance by capturing their interplay. By merging features across kernel scales while considering inter-dependencies on dimensions, the module enables a holistic and efficient adaptable fusion of multi-scale features with different shapes and leads to richer feature representations.

\subsection{Deep Large Kernel Pyramid Pooling Module}
\label{ssec:DLKPPM}

\begin{figure}[!t]
\centering
\includegraphics[width=\linewidth]{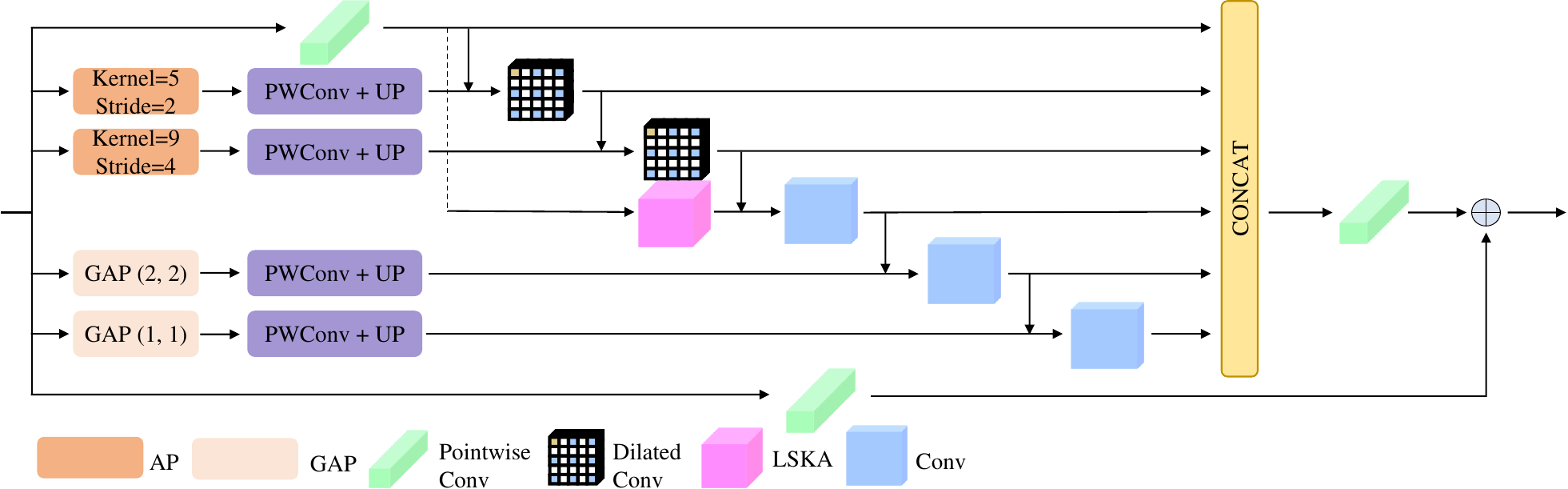}
\caption{\textbf{The structure of DLKPPM.} DLKPPM fuses multi-scale information using hierarchical kernels of various depths and sizes, minimizes the information loss from pooling with LKA, and uses dilated convolutions for small kernels.}
\label{fig:DLKPPM}
\end{figure}

DLKPPM preserves the hierarchical-residual structure and integrates kernels across various scales and depths, as illustrated in Fig.~\ref{fig:DLKPPM}, inspired by DAPPM~\cite{hong2021deep}.
To mitigate the loss of spatial details typically induced by large strides and pooling operations in conventional pyramid modules, DLKPPM uniquely integrates dilation convolutions and LKA mechanism with LSKA~\cite{lau2024large}, expanding the receptive field to 35. This efficient approach refines features and enhances semantic concept capture, while preserving crucial spatial information.

% DLKPPM preserves spatial details using hierarchical kernels and dilated convolutions to enrich contextual information while mitigating information loss from pooling.

\begin{figure}[!t]
\centering
\includegraphics[width=\linewidth]{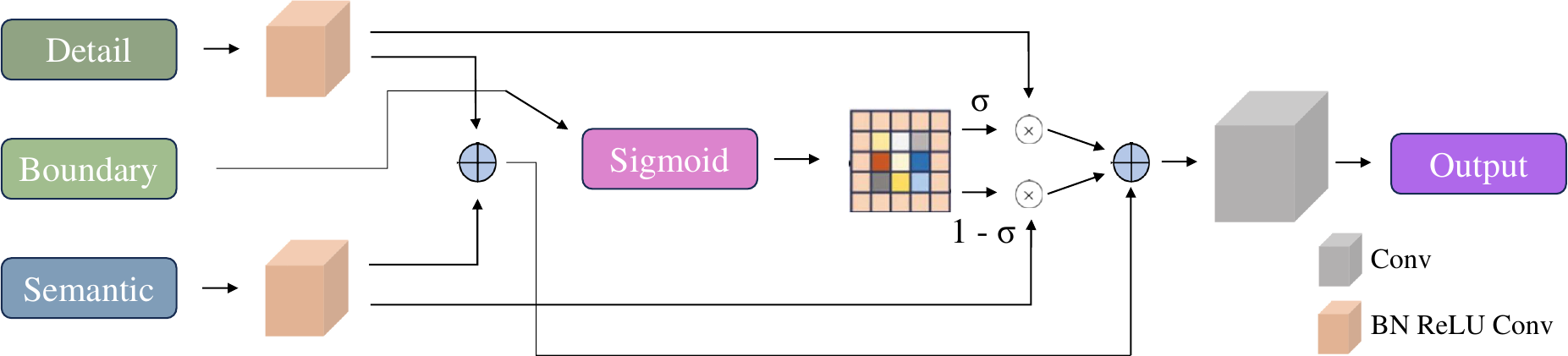}
\caption{\textbf{The structure of BGAF module.}
Semantic and detail information are refined through BN, ReLU, and convolution, while the boundary feature employs Sigmoid activation to compute a balancing weight $\sigma$ for adaptive merging. The balanced feature and shortcut are combined element-wise and processed through a final convolution to generate the output.}
\label{fig:BGAF}
\end{figure}

\subsection{Boundary Guided Adaptive Fusion}
\label{ssec:BGAF}

% The BGAF module provides advanced efficient multi-branch aggregation that balances contextual and spatial features with boundary information. As shown in Fig.~\ref{fig:BGAF}, it employs a shortcut residual connection to preserve critical feature information, thereby avoiding the degradation associated with a simple weighted summation. The module dynamically fuses high-level and low-level features, integrating robust semantic understanding with precise contour details. It further mitigates the limitations of the context branch's spatial precision and the detail branch's shallow semantic representation by adaptively adjusting feature contributions based on boundary significance. This approach ensures the precise detection of complete object boundaries and small fine-grained structures.
BGAF provides efficient multi-branch aggregation, balancing contextual and spatial features with boundary information. As shown in Fig.~\ref{fig:BGAF}, it employs a shortcut residual connection to preserve critical feature information, thereby avoiding degradation from simple weighted summation. The module dynamically fuses high-level and low-level features, integrating robust semantic understanding with precise contour details. It further mitigates the limitations of the context branch's spatial precision and the detail branch's shallow semantic representation by adaptively adjusting feature contributions based on boundary significance. This design ensures precise detection of complete object boundaries and fine-grained structures.

%The BGAF module is an efficient multi-branch aggregation framework that balances contextual and spatial features with boundary information. Fig.~\ref{fig:BGAF} depicts the details. It uses a shortcut connection to preserve critical feature information, avoiding degradation from simple weighted summation. The module dynamically fuses high-level and low-level features, integrating semantic understanding with precise contour details, and mitigates the limitations of the context branch's spatial precision and the detail branch's shallow semantic representation by adaptively adjusting feature contributions based on boundary significance. This ensures effective detection around object boundaries and small structures. 

% BGAF adapts semantic and spatial fusion using boundary information. It leverages shortcut connections and dynamic weighting to enhance object boundaries and small structures.

% !TeX root = ../main.tex

\begin{table*}[t]
  \centering

  \resizebox{0.72\linewidth}{!}{
  \begin{tabular}{lccccccc}
    \toprule
    Model & Resolusion & GPU & FPS $\uparrow$ & \#GFLOPs $\downarrow$ & \#Params (M) $\downarrow$ & mIoU (\%) $\uparrow$ \\
    \midrule
    BiSeNet(Res18) \cite{yu2021bisenet} &  1536 × 768 & GTX 1080Ti & 65.5 & 55.3 & 49 & 74.8 \\
    BiSeNetV2-L \cite{yu2021bisenet} & 1024 × 512 & GTX 1080Ti & 47.3 & 118.5 & - & 75.8 \\ \hline
    STDC1-Seg75 \cite{fan2021rethinking} & 1536 × 768 & RTX 3090 & 74.8 & - & - & 74.5 \\
    STDC2-Seg75 \cite{fan2021rethinking} & 1536 × 768 & RTX 3090 & 58.2 & - & - & 77.0 \\ \hline
    PP-LiteSeg-T2 \cite{peng2022pp} & 1536 × 768 & RTX 3090 & 96.0 & - & - & 76.0 \\
    PP-LiteSeg-B2 \cite{peng2022pp} & 1536 × 768 & RTX 3090 & 68.2 & - & - & 78.2 \\ \hline
    SFNet(DF2) \cite{li2020semantic} & 2048 × 1024 & RTX 3090 & 87.6 & - & 10.53 & 77.8 \\
    SFNet(ResNet-18) \cite{li2020semantic} & 2048 × 1024 & RTX 3090 & 30.4 & 247.0 & 12.87 & 78.9 \\ \hline
    DDRNet-23-S \cite{hong2021deep} & 2048 × 1024 & RTX 3090 & \textbf{108.1} & \textbf{36.3} & \textbf{5.7} & 77.8 \\
    DDRNet-23 \cite{hong2021deep} & 2048 × 1024 & RTX 3090 & 51.4 & 143.1 & 20.1 & 79.5 \\ \hline
    PIDNet-S-Simple \cite{xu2023pidnet} & 2048 × 1024 & RTX 3090 & 100.8 & 46.3 & 7.6 & 78.8 \\
    PIDNet-S \cite{xu2023pidnet} & 2048 × 1024 & RTX 3090 & 93.2 & 47.6 & 7.6 & 78.8 \\
    PIDNet-M \cite{xu2023pidnet} & 2048 × 1024 & RTX 3090 & 39.8 & 197.4 & 34.4 & 80.1 \\ \hline
    BEVANet (Ours) & 2048 × 1024 & RTX 3090 & 32.8 & 238.2 & 58.6 & \textbf{81.0} \\ 
    \bottomrule
  \end{tabular}
  }
  
  \caption{\textbf{Overall Quantitative Comparisons on Cityscapes~\cite{cordts2016cityscapes} Validation Set.} The bold numbers, indicating the best performance, emphasize the overall superiority of BEVANet. Most of the results are adopted from PIDNet~\cite{xu2023pidnet}.}
  \label{tab:overall comparison}
\end{table*}

\begin{table}[!ht]
  \centering
  
 %  \begin{tabular}{lccc}
 %    \toprule
 %    Model & GPU & FPS $\uparrow$ & mIoU (\%) $\uparrow$ \\
 %    \midrule
 %    PP-LiteSeg-T \cite{peng2022pp} & GTX 1080Ti
 % & 154.8 & 75.0 \\ \hline
 %    BiSeNetV2 \cite{yu2021bisenet} & GTX 1080Ti & 124.0 & 76.7 \\
 %    BiSeNetV2-L \cite{yu2021bisenet} & GTX 1080Ti & 33.0 & 78.5 \\ \hline
 %    DDRNet-23-S \cite{hong2021deep} & RTX 3090 & \textbf{182.4} & 78.6 \\
 %    DDRNet-23 \cite{hong2021deep} & RTX 3090 & 116.8 & 80.6 \\ \hline
 %    PIDNet-S \cite{xu2023pidnet} & RTX 3090 & 153.7 & 80.1 \\
 %    PIDNet-S-Wider \cite{xu2023pidnet} & RTX 3090 & 85.6 & 82.0 \\ \hline
 %    BEVANet-S (Ours) & RTX 3090 & 79.4 & \textbf{83.1} \\ 
 %    \bottomrule
 %  \end{tabular}
  
  \resizebox{1.0\linewidth}{!}{
  \begin{tabular}{lcccc}
    \toprule
    Model & GPU & FPS $\uparrow$ & \#GFLOPs $\downarrow$ & mIoU (\%) $\uparrow$ \\
    \midrule
    PP-LiteSeg-T \cite{peng2022pp} & GTX 1080Ti & 154.8 & - & 75.0 \\ \hline
    BiSeNetV2 \cite{yu2021bisenet} & GTX 1080Ti & 124.0 & - & 76.7 \\
    BiSeNetV2-L \cite{yu2021bisenet} & GTX 1080Ti & 33.0 & - & 78.5 \\ \hline
    DDRNet-23-S \cite{hong2021deep} & RTX 3090 & \textbf{182.4} & - & 78.6 \\
    DDRNet-23 \cite{hong2021deep} & RTX 3090 & 116.8 & - & 80.6 \\ \hline
    PIDNet-S \cite{xu2023pidnet} & RTX 3090 & 153.7 & 15.8 & 80.1 \\
    PIDNet-S-Wider \cite{xu2023pidnet} & RTX 3090 & 85.6 & 59.1 & 82.0 \\ \hline
    BEVANet-S (Ours) & RTX 3090 & 79.4 & 20.1 & \textbf{83.1} \\ 
    \bottomrule
  \end{tabular}
  }

  \caption{\textbf{Quantitative Comparisons on CamVid~\cite{brostow2009semantic}.} 
  %The bold numbers, indicating the best performance, emphasize the overall superiority of BEVAN. 
  Most of the results are adopted from PIDNet~\cite{xu2023pidnet}.}
  \label{tab:camvid comparison}
\end{table}

\section{Experiments}
\label{sec:exp}

\subsection{The Datasets and Implementation Details}
\label{ssec:dataset}

We mainly evaluated on Cityscapes~\cite{cordts2016cityscapes}, a widely recognized benchmark dataset for urban scene parsing, containing 2,975 training, 500 validation, and 1,525 testing images with a high resolution of $2048 \times 1024$. It includes 19 classes for semantic segmentation evaluation. We also use the CamVid~\cite{brostow2009semantic} dataset, consisting of 701 driving scene images at $960 \times 720$ resolution. It selects 11 classes of 32 annotated categories. 

%\subsection{Implementation Details}
%\label{ssec:implement}

% The models are pretrained on ImageNet \cite{russakovsky2015imagenet} with data augmentation like random cropping to 224×224 and horizontal flipping, following previous works \cite{he2016deep, hong2021deep, xu2023pidnet}. It lasts 100 epochs with a batch size of 256, using the SGD optimizer, an initial learning rate of 0.1 (reduced every 30 epochs), weight decay of 0.0001, and momentum of 0.9. 
% Main training mirrors previous research \cite{yu2018bisenet, wang2020deep, fan2021rethinking, hong2021deep, xu2023pidnet}, using the SGD optimizer with an initial learning rate of 0.01, a momentum of 0.9, and a weight decay of 0.001. A poly decay policy adjusts the learning rate, and data augmentation includes random cropping to 1024×1024, horizontal flipping, and scaling from 0.5 to 2.0. It runs for 484 epochs with a batch size of 10, incorporating Online Hard Example Mining (OHEM) for challenging sample selection. 
% Inference speed is measured on NVIDIA RTX 3090 GPU using PyTorch 2.4, CUDA 12.1, and Ubuntu 20.04, with a batch size of 1 for single-sample processing.

Our model is pretrained on ImageNet~\cite{russakovsky2015imagenet} using random cropping to $224 \times 224$ and horizontal flipping, consistent with prior works~\cite{yu2021bisenet, fan2021rethinking}. Pretraining runs for 100 epochs with a batch size of 256, employing SGD optimizer with a learning rate of 0.1, weight decay of 0.0001, and momentum of 0.9. 

% The main training phase, which follows similar protocols as in previous studies~\cite{yu2021bisenet, fan2021rethinking, hong2021deep, xu2023pidnet}, runs for 484 epochs with a batch size of 12 and learning rate of 0.008, utilizing a poly decay learning rate schedule, scaling from 0.5 to 2.0, and Online Hard Example Mining (OHEM). 
The main training phase, which follows similar protocols as in previous studies~\cite{yu2021bisenet, fan2021rethinking, hong2021deep, xu2023pidnet}, runs for 484 epochs with a batch size of 12 and a learning rate of 0.008 for Cityscapes, and 200 epochs with a batch size of 24 and a learning rate of 0.003 for CamVid~\cite{brostow2009semantic}. It employs a poly decay learning rate schedule (scaling from 0.5 to 2.0) and Online Hard Example Mining (OHEM) applied for challenging sample selection.

% Inference speed is measured on an NVIDIA RTX 3090 GPU using PyTorch 2.4, CUDA 12.1, and Ubuntu 20.04, with a batch size of 1 for single-sample processing.
All inference benchmarks were conducted on an NVIDIA RTX 3090 GPU using PyTorch 2.4, CUDA 12.1, and Ubuntu 20.04, with a batch size of 1 for single-sample processing.

\subsection{Quantitative Comparisons}
\label{ssec:compare}

\begin{table}[t]
  \centering
  \resizebox{0.85\linewidth}{!}{
  \begin{tabular}{lccc}
    \toprule
    Model & FPS $\uparrow$ & \#GFLOPs $\downarrow$ & mIoU (\%) $\uparrow$ \\ 
    \midrule
    PIDNet-S \cite{xu2023pidnet} & \textbf{93.2} & \textbf{47.6} & 76.32 \\
    PIDNet-M \cite{xu2023pidnet} & 39.8 & 197.4 & 78.22 \\
    PIDNet-L \cite{xu2023pidnet} & 31.1 & 275.8 & 78.25 \\ \hline
    BEVANet (Ours) & 32.9 & 238.2 & \textbf{79.27} \\
    \bottomrule
  \end{tabular}
  }
  \caption{\textbf{Quantitative Comparisons of Model Performance without Pretraining.} Ours outperforms PIDNet \cite{xu2023pidnet}.}
  \label{tab:comparison_without_pretraining}
\end{table}

%\subsubsection{Comparisons with pre-training}
%\label{sssec:overall_comparison}
\heading{Comparisons with pre-training.}
In this setting, all models are pre-trained on ImageNet~\cite{russakovsky2015imagenet} to ensure a fair comparison. As shown in Table~\ref{tab:overall comparison}, our BEVANet achieves state-of-the-art (SoTA) performance on Cityscapes~\cite{cordts2016cityscapes} above the real-time threshold of 30 FPS. With 81\% mIoU, our model outperforms similarly scaled benchmark models by leveraging optimized attention mechanisms and adaptive feature fusion, thereby delivering high accuracy and effectiveness without compromising speed. Furthermore, while PIDNet-M~\cite{xu2023pidnet} gains 1.3\% mIoU over PIDNet-S~\cite{xu2023pidnet} at a cost of 53 FPS, BEVANet improves upon PIDNet-M~\cite{xu2023pidnet} by 0.9\% mIoU with only a 7 FPS reduction.
Table~\ref{tab:camvid comparison} shows small-scale BEVANet-S also reaches SoTA on CamVid~\cite{brostow2009semantic} with 20.1 GFLOPs, indicating scalability and suitability for edge applications.
Notably, BEVANet-S reduces by nearly 40 GFLOPs compared to PIDNet-S-Wider~\cite{xu2023pidnet}, yet still achieves 1.1\% higher mIoU, clearly demonstrating its superior efficiency and accuracy.

%After pretraining on ImageNet \cite{russakovsky2015imagenet} for a fair comparison, as shown in Table \ref{tab:overall comparison}, our BEVAN model consistently surpasses the real-time threshold of 30 FPS, achieving state-of-the-art performance on Cityscapes. \cite{cordts2016cityscapes}. Our model surpasses benchmark models on the same scale, achieving near 81\% mIoU through optimized attention mechanisms and adaptive feature fusion, ensuring high accuracy and efficiency without sacrificing processing speed.

%\subsubsection{Comparisons without pre-training}
%\label{sssec:compare_wo_pretrain}

\heading{Comparisons without pre-training.}
In this setting, we compare our model primarily against the SoTA model PIDNet~\cite{xu2023pidnet}. As shown in Table~\ref{tab:comparison_without_pretraining}, without pre-training on ImageNet~\cite{russakovsky2015imagenet}, BEVANet achieves 79.3\% mIoU, outperforming PIDNet-M~\cite{xu2023pidnet}. Furthermore, it surpasses PIDNet-L~\cite{xu2023pidnet} by 1\% mIoU while delivering higher FPS. 
These results highlight the model’s efficiency and reduced dependency on large-scale pre-training, suggesting its potential for real-time applications where computational resources and labeled data are limited.

%In the setting without pre-training, we primarily compare our model against the state-of-the-art PIDNet~\cite{xu2023pidnet}.
%, where it demonstrates superior performance.
%Table~\ref{tab:comparison_without_pretraining} shows that, without pre-training on ImageNet~\cite{russakovsky2015imagenet}, our model achieves 79.3\% mIoU, outperforming state-of-the-art models like PIDNet-M~\cite{xu2023pidnet} and PIDNet-L~\cite{xu2023pidnet}. This highlights our model's reduced reliance on large datasets and its efficiency, making it ideal for real-world applications with limited data and time constraints.

\subsection{Ablation Study}
\label{ssec:abl}

%\subsubsection{Large Kernel Attention - EVA Block and DLKPPM}
%\label{sssec:EVA}

\heading{Architecture Efficiency.} As shown in the first two rows of Table \ref{tab:abl overall}, our architecture demonstrates increased speed with only a marginal reduction in performance, showcasing the efficiency of its semantic concept and spatial detail branches.

\heading{Large Kernel Attention.}
Our EVA block integrates the SDLSKA and CKS modules. The SDLSKA module, which combines LSKA~\cite{lau2024large} and SLaK~\cite{liu2022more}, significantly outperforms its components by improving the attention of the large kernel.
As indicated by the orange section in Table~\ref{tab:abl overall}, this improvement boosts mIoU by 0.8\% to 78.6\% while effectively capturing global context through an expanded receptive field. 
The pink section in Table~\ref{tab:abl overall} indicates CKS achieves a 0.26\% mIoU gain over LSKNet~\cite{li2023large} with a marginal 0.5 FPS reduction, highlighting its efficient fusion of multi-scale kernels across varying shapes and its integration of small and strip dilation kernels to balance spatial and channel information.
Collectively, the EVA block improves mIoU by 1.1\% with only a minor decrease in speed, thereby preserving real-time performance.

\begin{table}[t]
  \centering
  \resizebox{\linewidth}{!}{
      \begin{tabular}{llll|ccc}
        \toprule
        Architecture & Block & Selection Kernel & Branch Fusion & FPS $\uparrow$ & mIoU (\%) $\uparrow$ \\
        \midrule
        \cellcolor{yellow!30}PIDNet \cite{xu2023pidnet} & Convs & - & Bag \cite{xu2023pidnet} & 42.64 & 78.22 \\ 
        \cellcolor{yellow!30}BA (Ours) & \cellcolor{orange!30}Convs & - & Bag \cite{xu2023pidnet} & \textbf{44.25} & 77.77 \\ 
        BA (Ours) & \cellcolor{orange!30}SLaK\cite{liu2022more} & - & Bag \cite{xu2023pidnet} & 37.76 & 77.84 \\
        BA (Ours) & \cellcolor{orange!30}LSKA\cite{lau2024large} & -  & Bag \cite{xu2023pidnet} & 41.16 & 77.93 \\
        BA (Ours) & \cellcolor{orange!30}SDLSKA (Ours) & \cellcolor{pink!30}Addition & Bag \cite{xu2023pidnet} & 37.79 & 78.60 \\
        BA (Ours) & SDLSKA (Ours) & \cellcolor{pink!30}LSKNet \cite{li2023large} & Bag \cite{xu2023pidnet} & 37.46 & 78.72 \\
        BA (Ours) & SDLSKA (Ours) & \cellcolor{pink!30}CKS (Ours) & \cellcolor{purple!30}Bag \cite{xu2023pidnet} & 37.29 & 78.86 \\ 
        BA (Ours) & SDLSKA (Ours) & CKS (Ours) & \cellcolor{purple!30}Light\_Bag \cite{xu2023pidnet} & 38.48 & 78.39 \\ 
        BA (Ours) & SDLSKA (Ours) & CKS (Ours) & \cellcolor{purple!30}BGAF (Ours) & 32.85 & \textbf{79.27} \\
        \bottomrule
      \end{tabular}
      }
  \caption{\textbf{The ablation study comparisons of our modules without pretraining.} Each color represents a different module in the ablation study, with our module located at the bottom of each colored area.}
  \label{tab:abl overall}
\end{table}

\begin{table}[t]
  \centering
  \resizebox{0.72\linewidth}{!}{
  \begin{tabular}{lccc}
    \toprule
    PPM & FPS $\uparrow$ & mIoU (\%) $\uparrow$ \\
    \midrule
    DAPPM \cite{hong2021deep} & 32.85 & 80.43 \\ 
    PAPPM \cite{xu2023pidnet} & \textbf{33.38} & 79.97 \\ 
    DLKAPPM (Ours) & 32.47 & \textbf{80.96} \\ 
    \bottomrule
  \end{tabular}
  }
  \caption{\textbf{The ablation study comparisons of PPM.}}
  %\vspace{-5mm}
  \label{tab:abl PPM}
\end{table}

%\subsubsection{Branch Fusion}
%\label{sssec:BF}

\heading{Branch Fusion.}
Our BGAF module outperforms BAG~\cite{xu2023pidnet} by 0.4\% mIoU, as highlighted in the purple area of Table~\ref{tab:abl overall}. It enhances feature representation with shortcut paths and adaptive merging low- and high-level features with boundary information, emphasizing the value of balanced fusion and shortcuts.

%\subsubsection{Overall without pretraining}
%\label{sssec:overall}

\heading{Performance Enhancement.} In the second and last rows in Table~\ref{tab:abl overall}, under 30 FPS requirement, these innovations boost mIoU by 1.5\% to 79.3\%, demonstrating its effectiveness.

\heading{Multi-scale Fusion.}
For the first and last rows in Table \ref{tab:abl PPM}, our DLKPPM improves contextual fusion, providing a richer context and boosting mIoU by 0.5\% with a minimal 0.4 FPS reduction. 
Its larger receptive field reduces pooling information loss, making it well-suited for real-time applications.

\subsection{Qualitative Analysis}
\label{ssec:viz}

%\subsubsection{Small Object}
%\label{ssec:small_obj}
\begin{figure}[t]
\begin{minipage}[b]{\linewidth}
  \centering
  \centerline{\includegraphics[width=\linewidth]{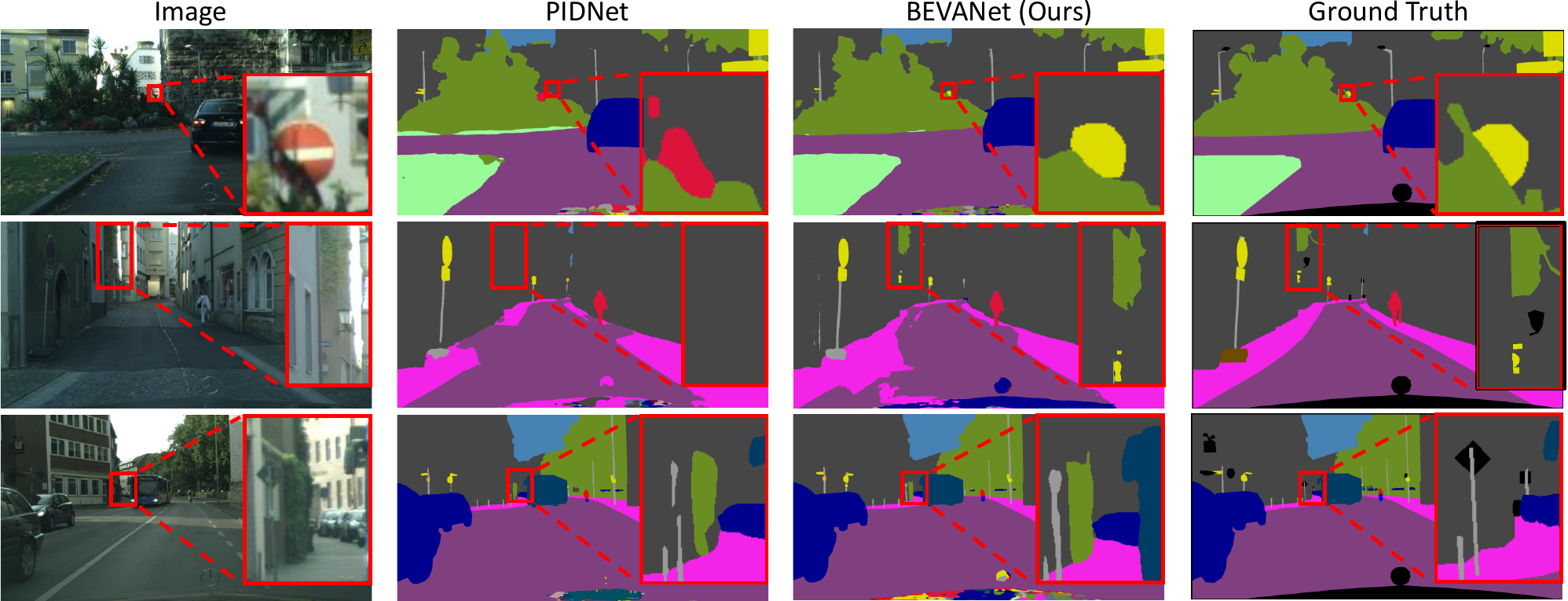}}
  \centerline{(a) The visualization comparison for the small objects. }\smallskip
\end{minipage}
\hfill
\begin{minipage}[b]{\linewidth}
  \centering
  \centerline{\includegraphics[width=\linewidth]{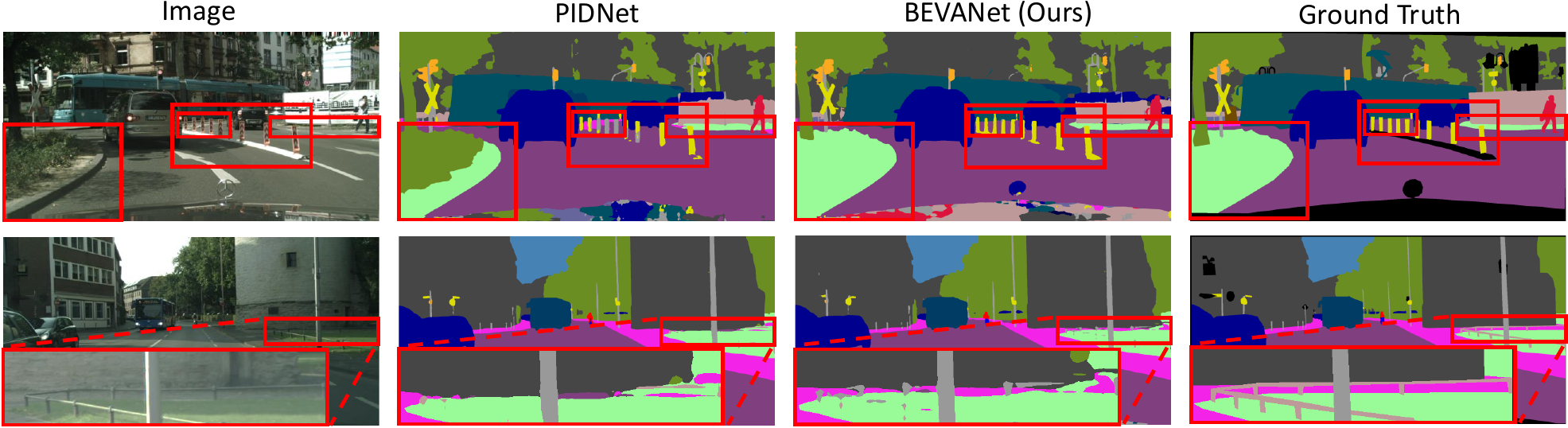}}
  \centerline{(b) The visualization comparison for the completeness.}
\end{minipage}
\caption{\textbf{Visualization comparisons.} Red denotes people, yellow indicates traffic signs, and green represents vegetation. From left to right, the columns display the image, PIDNet~\cite{xu2023pidnet} (baseline), BEVANet (ours), and Ground Truth, with each row corresponding to a different instance.}
\label{fig:viz}
\vspace{-1.5mm}
\end{figure}

\heading{Small Object.}
Our BEVANet outperforms PIDNet~\cite{xu2023pidnet} in detecting small objects, which are inherently difficult to identify. As depicted in Fig.~\ref{fig:viz}(a), it rectifies misclassifications such as confusing traffic signs for people, and detects plants and traffic signs that PIDNet~\cite{xu2023pidnet} misses. 
% BEVANet reliably identifies objects like vegetation even when they are unlabeled, 
BEVANet demonstrates the capability to identify unlabeled objects such as vegetation, indicating robust semantic understanding, accurate predictions, and a deeper comprehension of contextual information.

%\subsubsection{Completeness}
%\label{ssec:completeness}

\heading{Completeness.}
In Fig.~\ref{fig:viz}(b), our BEVANet leverages a large receptive field for thorough and accurate object detection, outperforming PIDNet in completely capturing traffic cones and grass. BEVANet also excels in handling larger objects and reliably detects sidewalks, where PIDNet~\cite{xu2023pidnet} fails, demonstrating its superiority in capturing reliable spatial information.
% !TeX root = ../main.tex

\section{Conclusion}
\label{sec:conclusion}

Our BEVANet model achieves competitive performance compared to state-of-the-art methods while reaching real-time processing at 33 FPS. Its key features, including the SDLSKA block for expanding receptive fields and the CKS mechanism for dynamic adjustments, enable accurate small object detection and refined boundaries. The bilateral architecture efficiently communicates between feature levels, and the BGAF module further enhances feature fusion. Additionally, our DLKPPM enriches feature representations. Future research will aim to optimize fusion strategies and reduce computational overhead to develop a lightweight large kernel attention model to further improve BEVANet's efficiency. 

% BEVAN introduces an efficient architecture for real-time semantic segmentation, achieving competitive accuracy and speed. Its innovative modules like SDLSKA and BGAF enhance both contextual understanding and boundary precision. Future work will focus on further reducing computational complexity.

%{\scriptsize \heading{Acknowledgements.} Thanks to NSTC, ASUSTek, and NTU for their support through grants 113-2634-F-002-007, 113-2221-E-002-112-MY3, 113L9009 and 114L892203.}

% Below is an example of how to insert images. Delete the ``\vspace'' line,
% uncomment the preceding line ``\centerline...'' and replace ``imageX.ps''
% with a suitable PostScript file name.
% -------------------------------------------------------------------------

%\vfill
%\pagebreak

% \input{contents/Copyright}
% \input{contents/Related Work Template}

\vfill\pagebreak

\bibliographystyle{IEEEbib}
% \bibliography{back/strings,back/references template}
\bibliography{back/references}

\end{document}